# Defining the Collective Intelligence Supply Chain

Iain Barclay, Alun Preece, Ian Taylor

Crime and Security Research Institute / School of Computer Science & Informatics

Cardiff University, UK

Email: {BarclayIS,PreeceAD,TaylorIJ1}@cardiff.ac.uk

**Abstract**

Organisations are increasingly open to scrutiny, and need to be able to prove that they operate in a fair and ethical way. Accountability should extend to the production and use of the data and knowledge assets used in AI systems, as it would for any raw material or process used in production of physical goods. This paper considers collective intelligence, comprising data and knowledge generated by crowd-sourced workforces, which can be used as core components of AI systems. A proposal is made for the development of a supply chain model for tracking the creation and use of crowdsourced collective intelligence assets, with a blockchain based decentralised architecture identified as an appropriate means of providing validation, accountability and fairness.

## Introduction

This paper presents an approach to artificial intelligence data provenance, with a focus on the use of a crowd-sourced workforce to derive collective intelligence in the form of digital assets or data. Making effective use of collective intelligence and crowd-sourced data presents opportunities for government and public sector organisations to better understand facts and experiences, improve decision making, produce new options and ideas, and give better oversight and monitoring of their actions (Saunders and Mulgan, 2017). We outline the challenges that organisations face in ensuring that they meet acceptable standards for transparency and accountability when using crowdworkers, and in ensuring that the originators of the work have been treated fairly and ethically for the work they've produced. The paper draws upon the model of supply chains used in industry to track the provenance of goods from raw materials through to finished product, and identifies the value in a collective intelligence supply chain as a means to allow for transparency of the production process and accountability for the provenance of the digital knowledge or data assets.

The paper describes some of the challenges that need to be addressed and understood in order to develop an effective collective intelligence supply chain model, and proposes a decentralised blockchain based architecture as an appropriate means to provide the necessary validation of the crowd workforce, fairness and accountability. The paper concludes by considering AI data in a wider context, and suggests that the proposed supply chain model could be extended to provide transparency and traceability of data from other sources, such as sensors, and licensed or open data.

## Motivation

Collective intelligence can be derived from large, loosely organised groups of people working together, such that the crowd provides more intelligence than the sum of the capabilities of the individuals (Malone, Laubacher and Dellarocas, 2010; Mulgan, 2017). Evidenced through the "commons based peer-production" (Benkler, 2006) of open source software, online community resources and citizen science projects[1], collective intelligence output can emerge from the apparent generosity of volunteers, or be driven by more tangible financial incentives through commercial crowdsourcing platforms, such as Figure Eight[2] or Amazon Mechanical Turk[3].

Crowdsourcing is often used to assemble groups of people to form a workforce for generating a body of Collective Intelligence, enabling the scale and skill-set of the resource to be readily tuned to changing organisational needs. A crowdsourced workforce can provide valuable human resources for training AI systems and operating as part of hybrid AI-human teams (Vaughan, 2018). Some crowdwork deployments require large numbers of unskilled workers to perform simple tasks (Alkhatib, Bernstein and Levi, 2017), whereas other deployments may require appropriately skilled or qualified workers to team up with AI systems (Kamar, Hacker and Horvitz, 2012; Scekic et al, 2015).

---

[1] https://www.zooniverse.org
[2] https://www.figure-eight.com/contributor/
[3] https://www.mturk.com

Organisations are increasingly open to scrutiny, and need to be able to prove that they operate in a fair and ethical way. In order to achieve and retain confidence and trust, organisations using AI systems need to be able to show the provenance and authenticity of the data and knowledge they use to make decisions (Diakopoulos, 2016, Doshi-Velez at al, 2017; Weitzner et al, 2008). When considering systems where a crowdsourced workforce has been used to create data assets, or where members of a crowd are used in a hybrid human-machine solution, this accountability should encompass the members of the crowdsourced workforce who created the assets or knowledge. In many cases, the organisation needs to be able to validate the identity and qualifications of the members of the workforce, so that they can show that the crowd they assembled were an appropriately diverse mix of genuine individuals, with appropriate skills, experience and qualifications for the task they undertook. For example, a medical system which required a body of trained medics (Hsu, 2017) would need to prove that the crowd were indeed medically trained, or a news service might need to prove impartiality. Without such assurances, the organisation could be unwittingly subject to bad actors - e.g., Sybil attacks (Wang et al, 2016; Miao et al, 2018) - maliciously trying to influence results, or non-experts interfering in systems where expertise is required.

From the other side of the crowdworking relationship, the organisation should offer the crowdsourced workers fair and ethical conditions for the work being done (Silberman, Irani and Ross, 2010). Understanding and designing systems to respond to the individual's motivations for becoming involved on a task (Crowston and Fagnot, 2008; Cuel et al, 2011), either on a voluntary or on a remuneration basis, will lead to improved satisfaction from the workforce, and improve the efficiency of the task by reducing dropouts (Eveleigh at al, 2014; Jackson et al, 2015). In terms of providing a fair reward to workers for producing assets, a survey of crowdworkers on the mTurk platform indicated a willingness to trade some portion of their payment for a future royalty-based payment on the use of the knowledge systems that they had worked on (Sriraman, Bragg and Kulkarni, 2017). Such a system would need to offer a way to track usage of the data or the trained model, so that royalties could be fairly and reliably paid to the knowledge creators (Zanzotto, 2017) - this offers technical challenges, as well as opening up the opportunity to "spin out" cooperative organisations or other bodies with membership made up of the crowd who created the original assets (Scheider, 2018).

In industries, such as manufacturing, agriculture and food production, it has become standard practice to track a product's lifecycle from its origin as raw materials through to finished goods, or food from farm to fork, with the relationships and data flows between suppliers modelled and tracked using supply chains (Mentzer et al, 2001; Parmigiani, Klassen and Russo, 2011). Research has been conducted into data supply chains, recording the provenance of data and digital information and proposing a "fair trade certificate for data" (Groth, 2013). A collective intelligence supply chain needs to extend the work done in data provenance, as consideration also needs to be given to providing accountability and transparency into the processes of validating the identity and appropriate qualifications of the crowd of data providers, and being able to demonstrate that the assets were produced in a fair and ethical way.

Much like the supply chains in industry and food production, development and use of a supply chain model to provide an audit trail of collective intelligence assets in use in AI systems has the potential to offer benefits in providing accountability and transparency for organisations making use of these assets in their operations, and ensuring that the assets have been created and are used in an ethical way.

# A Collective Intelligence Supply Chain

In developing a model for a collective intelligence supply chain, the challenges that need to be addressed include validation of the crowd workforce, fair treatment of the workforce and an ability to audit the overall system to provide accountability and transparency.

## Validating the Crowd

An organisation using a crowd-sourced workforce to produce data or digital assets, or to operate as part of hybrid human-AI interactions, needs to have assurance that the members of the workforce are suitable for the task. If particular skills, experience or qualifications are needed, then the organisation needs to be able to ensure that these criteria are met. This might be simple if the required workforce is small, but any validation scheme needs to be able to scale to an appropriate size - for example, a medical diagnosis system requiring thousands of qualified medics to perform collective diagnoses would need to have an efficient and cost-effective way to check the stated qualifications. Similarly, systems could rely on the collective skills and expertise of large numbers of civil engineers, firefighters, etc. The crowd of experts would be well placed to validate new members and add them to the growing pool, if they could be assured of genuine identity, though there is still a need to bootstrap the initial crowd.

Knowing that the stated identity of the members of the crowd is genuine would help to protect the organisation from disingenuous workers, who might have motivations to harm or disrupt operations. It would also enable a crowd of workers with different backgrounds to be assembled, ensuring a good diversity mix.

## Treating the Crowd Fairly

One of the shortcomings of paid crowdwork is that it is perceived as being unfair and exploitative to workers, with the work often compared to piece-work from bygone days (Kittur et al, 2013). Citizen science and other projects where work is voluntary provide contributors with intrinsic rewards, such as a feeling of making a contribution or gaining skills that can be used elsewhere, such as to enhance a career. The anticipated increase in the use of AI and collective intelligence should open new opportunities for paid and voluntary work to a global workforce, with a range of needs and motivations. The workers who make up this new workforce must be seen to be treated fairly and ethically, with the opportunity to receive motivating work with a fair and transparent reward, paid promptly. There are interesting opportunities to provide an ongoing usage-based reward for work done that gains adoption, rather than a simple one-time payment.

## Providing Accountability

Organisations which employ a crowd workforce need to be able to show that their workers have been treated fairly, just as any manufacturer of physical goods needs to be able to prove that they have not exploited the workers employed at any stage in the lifecycle of their products.

An organisation using data and AI models, potentially to make decisions with significant impact, needs to be able to show the provenance of the data that they use. This includes the origins of the data, demonstrating that the creators of the data were appropriately skilled or qualified, and not motivated by malicious intent.

A trail of data provenance should be immutable and non-repudiable, such that an audit can be honestly produced and not manipulated or denied by stakeholders when challenged. Organisations' asset procurement processes need to be transparent, trustworthy and to be able to be shown to meet accepted ethical standards.

# A Proposed Technology Platform Solution

The authors' experience in using blockchain technology to provide a secure messaging instructure with strong transparency and non-repudiation attributes (Barclay, 2017) suggests that blockchain technology could provide a base for development of a platform to meet the needs of a collective intelligence supply chain.

Solutions based on blockchain technology offer promise in terms of maintaining an honest record of an individual's identity, credentials, and reputation (Zyskind and Nathan, 2015; Tapscott and Tapscott, 2017), which would help meet the goal of validating the crowd - knowing that the individual members of the workforce were who they claimed to be, that they were suitably qualified for the task, and creating an environment where diversity of the workforce could be achieved.

In this context, there is work ongoing in the blockchain community in the area of identity, with projects from companies such as uPort[4] and Civic[5] allowing individuals to maintain a certified identity on a public blockchain. Validating that a user is a qualified doctor, or an experienced civil engineer requires checks on qualifications or evidence of experience, which is being tackled by a company called Skillschain[6] who are aiming to provide a verified and decentralised version of LinkedIn. Earning and keeping reputation from previous crowd work engagements can also be a helpful factor in crowd validation, with Colony[7] providing some work in this area.

Token Curated Registries (Goldin, 2017) provide a model to use economic incentives to moderate decentralised lists or directories without requiring a centralised owner or controller. This technique has the potential to support a scalable, self-regulating register of qualified members of a crowd-sourced workforce, such as doctors or engineers, and are in use in projects such as DAOStack[8], which is being designed to support communities and teams with membership in the hundreds of thousands.

---

[4] https://www.uport.me
[5] https://www.civic.com
[6] https://skillchain.io
[7] https://colony.io

The decentralised nature of a blockchain and the capabilities of programmable smart contracts could be used to escrow payment to ensure that workers are paid when work is done, and that they are not beholden to a central authority for payment or the risk of a payment platform shutting down before work is paid. Positive reputation can be earned for good work, and a permanent record of this could be stored with the user's credentials, enabling them to build a portfolio of experience on projects, with evidence that can be used to increase their suitability for other projects, or to prove their ownership in any ongoing equity models. Smart contracts can be used to design and enact autonomous organisations (Shermin, 2017), in which workers could automatically receive royalty-based income from their work on an ongoing basis, without needing to maintain an ongoing relationship with a potentially fragile, centralised authority.

Blockchain technology provides a decentralised distributed ledger of all transactions and the relationships between the parties involved in those transactions. The design of a public blockchain is such that it is prohibitively expensive to rewrite the transaction history, so it is well suited for storing a permanent record of exchanges between parties who do not have a trust relationship. The distributed ledger of blockchain provides a solid base for a supply chain (Kim and Laskowski, 2018), where transactions can be validated and the entire history of the product's creation be traced to provide provenance and assurance that appropriate standards have been met.

# Discussion and Conclusion

This paper has identified the need for a collective intelligence supply chain, which can be used to demonstrate that crowd-sourced data and knowledge assets used in AI systems have been produced by a validated workforce in a fair and ethical way. The supply chain provides transparency on the data's origins, and enables organisations to be held accountable for its quality and ethical production.

The authors have identified the challenges of ensuring the validity of the crowdsourced workforce, evidencing fairness and providing accountability that a collective intelligence supply chain needs to address, and propose that a platform built using decentralised blockchain-based technology is capable of meeting those challenges and supporting organisations in providing transparency and traceability on their data sources.

The challenge of demonstrating the validity of data sources used in AI systems, and showing that data is created and used fairly and ethically extends beyond collective intelligence and into the wider artificial intelligence domain. A system using data provided by remote sensors, for example, should require demonstrable assurance or certification that the sensors are correctly installed and calibrated, and that the data is not being maliciously tampered with enroute to the organisation's AI deployment. Where data is sourced externally to the organisation, either commercially or by using open data sources, it will generally be accompanied by licensing conditions governing the terms of use of the data, and any payment required for that use. Organisations should be able to demonstrate that their usage of the data in their AI systems is in accordance with the licensing conditions of the data, and that the correct payments and attributions are made for the use of the data.

Where data is shared among parties in a multi-agency coalition, assurance of the validity of the data source and traceability on the use of the data will need to extend outside the boundaries of a single organisation. In some sensitive cases, providing full visibility on the sources and uses of the data may not be appropriate, but the data will still need to be shown to be valid and trustworthy, and ethically sourced. Further, the owners of the data may only be willing to share the data if they can trace where, when and how their data has been used, even if that has to be obscured or abstracted in some way in sensitive cases.

The authors believe that there is scope to apply a supply chain model to the wider AI data lifecycle in order to provide appropriate levels of transparency and traceability for the data used, and will continue to develop their model to encompass crowdsourced data and data from a range of other sources, both internal to the organisation, externally sourced, and where sensitive data is shared among multi-agency parties.

---

[8] https://daostack.io

# References

Alkhatib, A., Bernstein, M.S. and Levi, M., 2017, May. Examining crowd work and gig work through the historical lens of piecework. In Proceedings of the 2017 CHI Conference on Human Factors in Computing Systems (pp. 4599-4616). ACM.

Barclay, I., 2017. Innovative Applications of Blockchain Technology in Crime and Security. MSc Dissertation, Cardiff University.

Benkler, Y., 2006. The wealth of networks: How social production transforms markets and freedom. Yale University Press.

Cuel, R., Morozova, O., Rohde, M., Simperl, E., Siorpaes, K., Tokarchuk, O., Wiedenhoefer, T., Yetim, F. and Zamarian, M., 2011. Motivation mechanisms for participation in human-driven semantic content creation. International Journal of Knowledge Engineering and Data Mining, 1(4), pp.331-349.

Crowston, K. and Fagnot, I., 2008, July. The motivational arc of massive virtual collaboration. In Proceedings of the IFIP WG 9.5 Working Conference on Virtuality and Society: Massive Virtual Communities (pp. 1-2).

Diakopoulos, N., 2016. Accountability in algorithmic decision making. Communications of the ACM, 59(2), pp.56-62.

Doshi-Velez, F., Kortz, M., Budish, R., Bavitz, C., Gershman, S., O'Brien, D., Schieber, S., Waldo, J., Weinberger, D. and Wood, A., 2017. Accountability of AI under the law: The role of explanation. arXiv preprint arXiv:1711.01134.

Eveleigh, A., Jennett, C., Blandford, A., Brohan, P. and Cox, A.L., 2014, April. Designing for dabblers and deterring drop-outs in citizen science. In Proceedings of the SIGCHI Conference on Human Factors in Computing Systems (pp. 2985-2994). ACM.

Goldin, M. 2017. Token-Curated Registries 1.0. Available at: https://docs.google.com/document/d/1BWWC__-Kmso9b7yCI_R7ysoGFIT9D_sfjH3axQsmB6E/edit [Accessed: 19 July 2018].

Groth, P., 2013. Transparency and reliability in the data supply chain. IEEE Internet Computing, 17(2), pp.69-71.

Hsu, J., 2017, August. "Can a Crowdsourced AI Medical Diagnosis App Outperform Your Doctor?", Scientific American. Available at: https://www.scientificamerican.com/article/can-a-crowdsourced-ai-medical-diagnosis-app-outperform-your-doctor/ [Accessed: 9 July 2018].

Jackson, C.B., Østerlund, C., Mugar, G., Hassman, K.D. and Crowston, K., 2015, January. Motivations for sustained participation in crowdsourcing: case studies of citizen science on the role of talk. In System Sciences (HICSS), 2015 48th Hawaii International Conference on (pp. 1624-1634). IEEE.

Kamar, E., Hacker, S. and Horvitz, E., 2012, June. Combining human and machine intelligence in large-scale crowdsourcing. In Proceedings of the 11th International Conference on Autonomous Agents and Multiagent Systems-Volume 1 (pp. 467-474). International Foundation for Autonomous Agents and Multiagent Systems.

Kim, H.M. and Laskowski, M., 2018. Toward an ontology‑driven blockchain design for supply‑chain provenance. Intelligent Systems in Accounting, Finance and Management, 25(1), pp.18-27.

Kittur, A., Nickerson, J.V., Bernstein, M., Gerber, E., Shaw, A., Zimmerman, J., Lease, M. and Horton, J., 2013, February. The future of crowd work. In Proceedings of the 2013 conference on Computer supported cooperative work (pp. 1301-1318). ACM.

Malone, T.W., Laubacher, R. and Dellarocas, C., 2010. The collective intelligence genome. *MIT Sloan Management Review*, *51*(3), p.21.

Mentzer, J.T., DeWitt, W., Keebler, J.S., Min, S., Nix, N.W., Smith, C.D. and Zacharia, Z.G., 2001. Defining supply chain management. Journal of Business logistics, 22(2), pp.1-25.

Miao, C., Li, Q., Su, L., Huai, M., Jiang, W. and Gao, J., 2018, April. Attack under Disguise: An Intelligent Data Poisoning Attack Mechanism in Crowdsourcing. In Proceedings of the 2018 World Wide Web Conference on World Wide Web (pp. 13-22). International World Wide Web Conferences Steering Committee.

Mulgan, G., 2017. Big Mind. Princeton University Press.

Parmigiani, A., Klassen, R.D. and Russo, M.V., 2011. Efficiency meets accountability:

Performance implications of supply chain configuration, control, and capabilities. Journal of operations management, 29(3), pp.212-223.

Saunders, T. and Mulgan, G., 2017. Governing with collective intelligence. Nesta, London. Available at: https://media.nesta.org.uk/documents/governing_with_collective_intelligence.pdf [Accessed: 28 August 2018].

Scekic, O., Miorandi, D., Schiavinotto, T., Diochnos, D.I., Hume, A., Chenu-Abente, R., Truong, H.L., Rovatsos, M., Carreras, I., Dustdar, S. and Giunchiglia, F., 2015, October. SmartSociety--A Platform for Collaborative People-Machine Computation. In Service-Oriented Computing and Applications (SOCA), 2015 IEEE 8th International Conference on (pp. 147-154). IEEE.

Schneider, N., 2018. An internet of ownership: democratic design for the online economy. The Sociological Review, 66(2), pp.320-340. Online: https://ioo.coop/2016/09/07/an-internet-of-ownership/

Shermin, V., 2017. Disrupting governance with blockchains and smart contracts. Strategic Change, 26 (5), pp.499-509.

Silberman, M., Irani, L. and Ross, J., 2010. Ethics and tactics of professional crowdwork. XRDS: Crossroads, The ACM Magazine for Students, 17(2), pp.39-43.

Sriraman, A., Bragg, J. and Kulkarni, A., 2017, February. Worker-owned cooperative models for training artificial intelligence. In *Companion of the 2017 ACM Conference on Computer Supported Cooperative Work and Social Computing*(pp. 311-314). ACM. Available at: https://homes.cs.washington.edu/~jbragg/files/sriraman-cscw17.pdf

Tapscott, D. and Tapscott, A., 2017. How blockchain will change organizations. MIT Sloan Management Review, 58(2), p.10.

Vaughan, J.W., 2018. Making Better Use of the Crowd: How Crowdsourcing Can Advance Machine Learning Research. Journal of Machine Learning Research, 18(193), pp.1-46.

Wang, G., Wang, B., Wang, T., Nika, A., Zheng, H. and Zhao, B.Y., 2016, June. Defending against sybil devices in crowdsourced mapping services. In Proceedings of the 14th Annual International Conference on Mobile Systems, Applications, and Services (pp. 179-191). ACM.

Weitzner, D.J., Abelson, H., Berners-Lee, T., Feigenbaum, J., Hendler, J. and Sussman, G.J., 2008. Information accountability. Communications of the ACM, 51(6), pp.82-87.

Zanzotto, F.M., 2017. Human-in-the-loop Artificial Intelligence. arXiv preprint arXiv:1710.08191.

Zyskind, G. and Nathan, O., 2015, May. Decentralizing privacy: Using blockchain to protect personal data. In Security and Privacy Workshops (SPW), 2015 IEEE (pp. 180-184). IEEE.